# ANALYSIS ON ADVANCEMENT OF HYBRID FUZZY SLIDING MODE CONTROLLERS FOR NONLINEAR SYSTEMS

Mukhtar Fatihu Hamza[1*], Abdulbasid Ismail Isa[2] and Jamilu Kamilu Adamu[3]

[1] *Department of Mechatronics Engineering, Bayero University, Kano.*
[2] *Department of Electrical and Electronics Engineering, Usmanu Danfodiyo University Sokoto.*
[3] *Department of Engineering Services, Federal Ministry of Power, Works and Housing.*
*Corresponding author mfhamza.mct@buk.edu.ng*

**Abstract:** Chattering phenomena is the major problem affecting sliding mode control (SMC). Also, finding a suitable structure and appropriate parameters values of fuzzy logic system (FLS) is a complex and difficult task. In addition, the stability of a general FLS is difficult to guarantee. Many types of combinations between FLS and SMC have been used to form an intelligent and robust controller that deviates from the limitations of each constituent and benefit from the advantages of each constituent. In this study, a survey of recent developments on the Hybridization of FLS (type-1) and SMC is presented. In addition, the differences between using the SMC in FLC or using FLC in SMC as well as their limitation and advantages are highlighted. It is found that the majority of the combinations made are intended to approximate the nonlinear sliding surface within the boundary layer. Limitations of the previous approaches and future research directions are pointed out. For novice researchers, this survey can serve as a foundation for their work while for expert researchers this review can serve as a benchmark for further advancement and exploration of other hybridization methods.

## 1. INTRODUCTION

It is widely known that sliding mode control (SMC) effectively provides robust control for nonlinear system even in the presence of uncertainties and disturbances [1]. This method has been applied successfully to many kinds of systems due to its attractive features [2]. These features include easy realization, compatibility to multiple input and multiple output (MIMO) systems, good control performance for nonlinear system such as good transient response, fast response and insensitivity to external disturbance and/or plant parameter variation [3, 4]. In addition, it is possible to guarantee the stability of SMC since it benefits from the merit of switching control law [1, 2]. However, the switching control law introduced chattering in the system. This is due to its alternating switching, which occurs from its digital implementation [4]. On the other hand fuzzy logic controller (FLC) inherits many attractive features which includes easy incorporation of expert knowledge into the control law, less model dependent, robust, easily used to model linguistic rules [5]. Yet, the stability of a general FLC is not easy to confirm [6]. Also, despite the importance of FLC, there is no organised and comprehensive procedure for the design of FLC [7].

Though, a FLC is shown to be similar to a SMC [8], the integration of FLC and SMC to form fuzzy sliding mode control (FSMC) has been attracting unprecedented interest from researchers [9-20]. This is because the FSMC provide an intelligent and robust control that benefits from the merits of both constituents while eliminate or reduce the inherent drawback of the constituents (FLC and SMC) [12]. For example, the ability of SMC to decouple higher dimensional systems into lower dimensional subsystems and its invariance property can be used to decrease the size of FLC rule base [2, 21-24]. In addition, FLC can effectively eradicate chattering effect of SMC via the replacement of crisp switching surfaces by fuzzy boundary layers [4, 25-27]. Moreover, FLC can be designed in the form of (fuzzy proportional integral derivative (PID) or proportional derivative (PD) or proportional integral (PI) control, conventional FLC, model based FLC) and include the supervisory SMC in order to improve the robust performance and guarantee the stability of the closed loop systems [28-36]. Furthermore, the FLC design issue and its stability analysis can be addressed within a SMC framework [1, 2, 4, 27, 37, 38].

In practice, the mathematical models for many complex systems are not available [4]. This attracts the attention of researchers to the adaptive control. Adaptive control means the control of the partially known plant using some adaptation mechanism [4]. Recently, most research in SMC and FSMC employed adaptive fuzzy system for tuning the parameters of SMC and FSMC [34, 39-43]. The most commonly used FSMC in literature are: FSMC based on equivalent control, SMC based on fuzzy switch-gain regulation, SMC based on fuzzy system approximation and adaptive FSMC [44-47].

The stability of FSMC systems can be analysed using Lyapunov stability theory. The common Lyapunov functions that can be used are: Global or common quadratic Lyapunov functions, piecewise quadratic Lyapunov functions and non-quadratic or fuzzy Lyapunov functions [4]. The detailed analysis and formulation of these Lyapunov functions is presented in [4, 48, 49]. There are many studies on the combination of FLC, SMC, PID, PD and PI, to design an adaptive, robust and intelligent controller. Despite the existence of these studies, we have not found a review of recent advances on this significant subject. Though, Piltan *et al*. [50] presented a review on design and implementation of sliding mode algorithm. However, the authors mainly focus on their applications to the robot manipulator and cover papers up to 2011 only. In addition, complicated systems such as power systems, synchronization of chaotic systems, motor systems, mechanical systems, and etc. are ignored in [49]. Yu and Okyay [51] made a survey on combination of SMC with soft computing techniques. The authors focus on the use of NN, FLS and probabilistic reasoning in SMC to alleviate the shortcomings of the conventional SMC methods. Furthermore, the review presented in [51] covered papers up to the year 2009.

In this study, we present a survey of the recent developments on the combination of FLC (type-1 fuzzy logic controller), SMC, PID, PD and PI to produce an adaptive, robust and intelligent controller. The review is conducted to summarize the state-of-the-art literature and point out unresolved problems before suggesting future research. To prevent repetition of the study that has been conducted. Additionally, we propose to offer a vivid viewpoint with a comprehensive and detailed survey of the studies in this field.

**2. FUZZY SLIDING MODE CONTROL**

The SMC and FLC are normally hybridized in two ways: (1) To use SMC in a fuzzy (FLC is a main controller), (2) To use FLC in a SMC (SMC is a main controller). Each of these combinations has its own advantage and disadvantage. In the first case, the control command *u* is approximated by FLC as a non-linear function of *s* in the boundary layer [52]. The resulting combination is actually a FLC with single input, single output (SISO) [53-55]. The *s* and *u* are the input and output of the resultant combination respectively. This type of combination has a typical rule as in the following format:

IF $s$ is NS and $\dot{s}$ is NS THEN $u$ is NS or a constant

The boundary layer of the system is effectively increasing by this rule format [27]. The fuzzification level of *s* and *u* determined the number of rules in the rule base. Though, the number of rules is much decreasing compared with a classical FLC. The fuzzification of s for this combination is illustrated in Figure 3. This combination produced a nonlinear boundary layer which is an advantage when compared with a conventional SMC (it have linear boundary layer). However, the form of boundary layer add little to the control performance in view of the fact that the width of the boundary layer is more influential to the control performance [1]. The main disadvantage of this combination is that the value of λ in equation (6) has to be predefined by an expert. Generally, this type of combination intended to approximate only a nonlinear sliding surface instead of nonlinear switching function. The switching function is still linear as in equation (6).

In the second case, the rule output function for a Takagi Segino King fuzzy type (TSK) is typically a linear function of controller inputs. The switching function of a SMC is similar to the mathematical expression of this function. This shows that the FLC can be design using the information from SMCr and the resulting controller is still a typical TSK type FLC. In view of the fact that FLS can connect different control algorithms into a single system without a glitch, one can take direct approach to incorporate SMC in to FLC [13, 56]. For example each rule in a FLS can be a SMC of different form. The coefficient of sliding surface and the boundary layer become the coefficients of the function of the output rule and have their physical implications. The number of input to the FLC is equal to the number of state variables. The rule base size does not reduce by the structure of the FLC. However, the switching function of this form of combination is more complicated with less rules compared with conventional FLS. The $i^{th}$ rule of this kind of combination can be express as follows [5, 57-59]:

IF $e$ is $A_i$ and $\dot{e}$ is $B_i$, THEN $u_i = k \tanh\left(\dfrac{\dot{e} + \lambda_i + c_i}{\varphi_i}\right)$

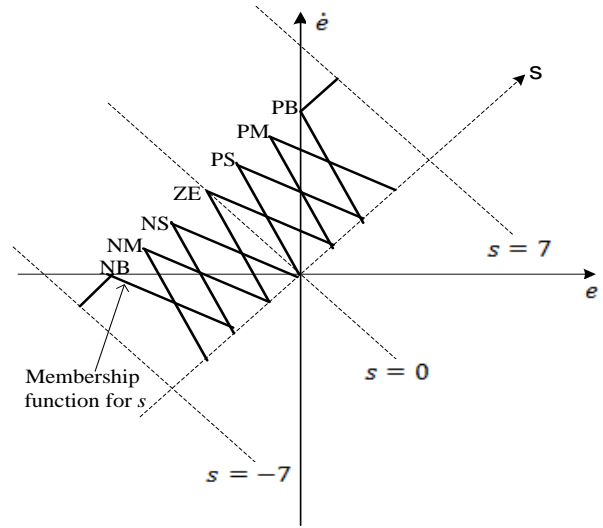

**Figure 3:** Fuzzification of $s$

Note that the rule output function can also be Sign or saturation functions. Figure 4 show the fuzzification of $\dot{e}$ and $e$. This type of combination can easily approximate the nonlinear switching function with just a few rules. In addition, for two-dimensional systems, the switching line is a function of $\dot{e}$ or $e$ which is one-dimensional function. Very small number of rules is required to approximate one-dimensional function. This is how this type of combination reduces the rule base size.

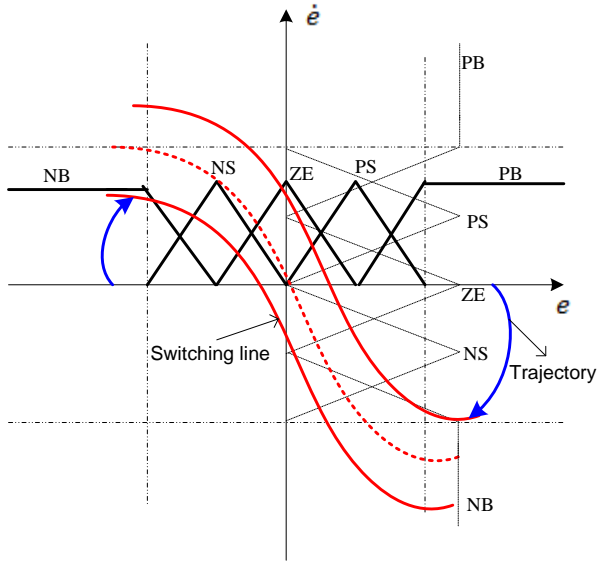

**Figure 4:** Fuzzification of $e$ and $\dot{e}$

### 3. FUZZY SLIDING MODE CONTROL

This section provides a concise overview on the combinations of FLS, SMC, PID, PD, and PI to produce an adaptive and robust intelligent controller.

There are many researches on hybridization of different kind of FLS with different form of SMC. in majority of these studies' success was achieved. This section is for the discussion of review of these literatures. The discussion demonstrates the effectiveness of using the different form of FSMC for different control applications. The literature that applied FSMC in real life experiments and simulation studies are highlighted. For example, Zhang and Guo [11] presented a novel universal function projection lag synchronization (UFPLS). The active SMCr (ASMC) is introduced for UFPLS chaotic systems. Integral sliding surface is used to determine the sliding parameters. To eliminate the chattering effect, the FLS is applied to adjust the coefficient of switch term. Several simulations are performed for fuzzy ASMC (FASMC) and ASMC. The proposed method is effective as shown by the obtained results. The SMC problem for a class of TSK FLS with matching uncertainties is addressees by Zhang *et al*. [10] they proposed a memory based sliding surface which is made up of the current state and the delayed state. Both adaptive and robust FSMC are designed based on the presented memory sliding surface. It is illustrated that the closed-loop control system is asymptotically stable and the sliding surface can be reached. Some continuous SMCs are presented to reduce the chattering. The simulation results obtained shows that the proposed memory based FSMC is more effective than the conventional memory less FSMC in terms of the stability guaranteed and transient performance.

Zhang and Wang [56] presents the FSMC for uncertain nonlinear systems based on TSK fuzzy model. Both inputs and state matrices have the parameter uncertainties and external disturbance exit. The proposed controller has the ability to deal with systems devoid of assuming the control matrices of each local TSK model to be similar. The condition for the survival of the proposed FSMC is found based on linear matrix inequality (LMI). The proposed method is effective as shown by the results obtained in numerical example. Taghian Dinani *et al*. [60] proposed simulation study for regulation of blood glucose concentration in type-1 diabetics. This is done by integrating the single order SMC and the FLS theory (FSOSMC). The gain of the proposed controller is tuned adaptively online. This reduces the amount of control effort and attenuates of the chattering effect of SMC. The simulation results obtained shows the effectiveness and robustness of the method against the parameter variation and meal disturbance rejection. Moreover, the result demonstrated the superiority of FSOSMC over fuzzy high order SMC (FHOSMC) proposed in [61]. In Arslan et al [62] a FLS hybridized with SMC is designed for vehicle vibration reduction. The human body is incorporated into a full vehicle model and considered as a lumped parameter model. The simulation results indicated that the proposed method is superior to the conventional SMC.

Ullah *et al*. [39] proposed an adaptive fuzzy fractional order SMC (FFOSMC) for a high performance servo actuation system that is subjected to aerodynamic loads and uncertainties. A servo position controller was proposed by employing the fractional calculus and verifying its performance under nonlinear friction, system uncertainties and aerodynamic loads. The authors utilized the merit of fractional order proportional-derivative PD$\lambda$ sliding surface and fractional order proportional-integral PI$\alpha$ sliding surface. The FLC can use small switching gain of the discontinuous control under large upper bounded uncertainties. The sliding condition is guaranteed by formulating the adaptive laws based on Lyapunov function. The simulation results demonstrate the advantage of the proposed controller compared to the conventional SMC and PID controllers.

Mokhar and Ismail [63] proposed the intelligent control (region tracking) of a single autonomous under water vehicle (AUV) using the gain scheduling FSMC. The Signum function is replaced by saturation function. The simulation result obtained shows the very good performance of the proposed method. In another work by Mokhar and Ismail [64] the same method in [63] was compared with adaptive FSMC for control of AUV. In this case, the reaching law (switching term) is modelled using FLS theory. Based on the simulation result obtained, it have been concluded that, gain scheduling FSMC have superior performance over adaptive FSMC in terms of robustness. Yoo *et al*. [65] study the intelligent control of Smart Unmanned Aerial Vehicle (SUAV) using FSMC. The SMC is employed to guarantee the stability of the system under varying loads and satisfy its target performance. FLC was used to for intelligent selection of the control switch. The actuator controller of the SUAV receives the control commands from the Digital Flight Control Computer based on the FSMC scheme, and transmits them to actuators under the feedback control. The effectiveness of the proposed method was verified using simulation test and the result shows the superiority of FSMC over PID controller.

Lakhekar and Waghmare [36] developed a novel dynamic FSMC algorithm for heading angle control of AUV. Single input Mamdani type fuzzy SMC was designed. Two supervisory FLS was designed for tuning of lifting gain and boundary layer of FSMC. The simulation results show that, the approaching phase tracking error along with chattering effect was drastically reduced. In addition, the minimum reaching time was achieved for output tracking response compared with conventional SMC. Nejatbakhsh *et al.* [28] Proposed enhanced time delay control (TDC) method for an Underwater Vehicle-Manipulator System (UVMS). Three different terms were used for the proposed controller: A Terminal Sliding Mode (TSM) term that gives a quick response, a time-delay-estimation term that cancels nonlinearities of the UVMS dynamics and a PID term that decreases the tracking error. Fuzzy rules were used to adaptively tune the gains of PID and TDC terms. Simulation results shows that the proposed controller gives the good performance in trajectory tracking and has a satisfactory robustness under the external disturbance and unknown torques /forces in comparison with conventional SMC

Lakhekar *et al.* [66] designed the enhanced dynamic FSMC (EDFSMC) to track periodic commands in vertical plane for AUV. Fuzzy adaptive tuner was used to improve the tracking performance, which is applied to shift input-output MF of FSMC algorithm. Two supervisory FLSs was used to tune the support of output singleton functions and to vary the width of boundary layer. The proposed method have the significant benefits as free from chattering, robust to uncertainties, simple control framework and stable tracking control performance. The system's stability is proven based on Lyapunov method. The considerable improvement is observed for tracking performance from the simulation results obtained compared with traditional SMC and FSMC. Khanesar *et al.* [13] study FSMC, where a FLS is employed to estimate the nonlinear dynamical system online, and the networked-induced delay is controlled by Pade approximation. The problem of packet losses is solved by viewing them as big time-varying delays in the system. Lyapunov function used confirms that the tracking error converges almost to zero asymptotically and derived the adaptation laws parameters which are found to be stable. Simulation results found demonstrate that the proposed controller is able to control nonlinear dynamical systems over a network, which is subject to time-varying network-induced delays, bounded external disturbances, and packet losses with adequate performance.

Filabi and Yaghoobi [42] presented the adaptive FSMC for trajectory control of 6 degree of freedom (DOF) parallel manipulator. The proposed controller consists of fuzzy approximator which estimates the nonlinear function of the plant and robustifying the control terms. Simulation results show that, the proposed method has achieved favourable control efficiency in respect to external disturbances, nonlinearities and uncertainties. Yoshimura [67] presented the design of an adaptive FSMC for uncertain discrete-time nonlinear dynamic systems. The uncertainties include the external disturbances and the modelling errors. FLS is used to approximate the nonlinear uncertainties based on the universal approximation theorem. An adaptive complementary term is added to the proposed adaptive FSMC to compensate approximation error. To reduce the number of rule, the extended single input rule modules are proposed. Weighted least squares estimator was used to estimates for the un-measurable states and the adjustable parameters. The results of the simulation experiment of a simple numerical system demonstrate the effectiveness of the proposed method. Pezeshki *et al.* [68] studied the control of an under actuated overhead crane system based on an adaptive FSMC using fuzzy approximation and a model-free adaptive controller using feedback linearization. The control process for both controllers is using load swing angle and trolley position. In adaptive FSMC, SMC is used adaptive TSK fuzzy algorithm to update estimation of unknown function. The simulation result shows the robustness of this method. Hasheminejad *et al.* [69] presented an adaptive FSMC scheme for actively suppress the two-dimensional vortex-induced vibrations of an elastically mounted circular cylinder. The proposed scheme comprises of a FLC designed to imitate an ideal SMC, and a robust controller in tended to compensate for the difference between the FLC and the ideal one. The FLC parameters and the uncertainty bound of the robust controller are adaptively tuned online. The simulation results indicate the superiority of the proposed scheme over PID controller.

Guo *et al.* [70] came up with a SMC according to exponential reaching law for the anti-lock braking system (ABS) to achieve best slip value. FLC was used to optimize the reaching law. Simulation results indicates the effectiveness of the proposed method compared with a conventional Bang-bang ABS controller. Noh and Choi [15] studied the design of an integral SMC for a nonlinear boost DC–DC converter based on the TSK fuzzy approach. The sliding surfaces are derived using LMIs. A switching feedback method was employed to guarantee the condition of reachability. The simulation and experimental results indicate that the proposed method can robustly regulate the output voltage in presence of bounded model uncertainties. Mon and Lin [71] investigated the robust non-linear FSMC application for a high-voltage direct current system based on voltage source converter. FLS is used to cancel non-linearity and the SMC offers invariant stability to modelling uncertainties pertaining of converter parameter. The simulation results of the proposed controller indicate its excellent transient and steady-state performance and complete decoupled control of active and reactive power. Mienet *et al.* [72] developed an output feedback tracking control method for robot manipulators with uncertainties and only position measurements. First a quasi-continuous second-order SMC (QC2C) was designed. Then an adaptive fuzzy QC2C (FQC2C) is designed to attenuate the chattering, in which the FLS was employed to adaptively tune the SMC gain. The controllers integrated in a super-twisting second-order sliding mode observer such that a robust exact differentiator can estimate the sliding manifold derivative and to estimating the joint velocities. The simulation results for a PUMA560 industrial robot verify the effectiveness of the proposed method.

Hsu *et al*. [73] studied an enhanced FSMC for position tracking control of a linear induction motor. The SMC is based on the back-stepping control method with the combination of two FLC. The dynamic tune of the sliding surface slope constant of the SMC was based on the controlled system states by the first FLC. The upper bounds of the lumped uncertainties were estimated using the fuzzy inference mechanism of the second FLC. Experimental result demonstrates the effectiveness and robustness of the presented scheme.

Gao *et al*. [17] studied the universal integral SMC problem (dynamic integral sliding mode control (DISMC) scheme) for the general stochastic nonlinear systems modelled by Itô type stochastic differential equations. The proposed DISMC method does not include the two very restrictive assumptions in most existing integral SMC schemes to stochastic FLS. The stochastic Lyapunov theory was used to shown that the closed-loop control system trajectories are reserved on the integral sliding surface approximately surely from the initial time. LMI was employed to guarantee the stochastic stability of the sliding motion. Simulation results illustrate the advantages and effectiveness of the proposed method.

Do *et al*. [74] described an adaptive FSMC for velocity control of secondary control units of secondary controlled hydrostatic transmission (SC-HST) system. The proposed method can deal with the disturbance load and nonlinearities of a SC-HST in real application. Lyapunov theory was used to derive the adaptive laws, thus the stability of the closed loop system is confirmed. Experimental results show the high performance of the presented method compared with the PID controller. Szabat *et al*. [29] presented a novel FSMC structure. The main controller is a fuzzy TSK type whose control surface is separated into local models (LM). In the LM located further from the equilibrium point the control lows are in form of conventional SMC. The division specified as an origin is described as a function of traditional PI controller. Due to the special location of the MF a soft switching between sectors is guaranteed. The theoretical considerations are proven by a simulation and experimental study.

## 4. GENERAL OVERVIEW, TREND, LIMITATIONS AND FUTURE RESEARCH

The general summary of this domain of research is presented in this section. Based on the review result of this domain, the probable future trends that can be visualized are discussed. The limitations of previous approaches and future research direction are outlined in the section.

### 4.1 Overview of the Area

It is shown that the hybrid sliding mode controller can solve complex control problems such as in robotics, power systems, synchronization of chaotic systems, motor systems, mechanical systems etc. Moreover, it can handle the stability analysis for uncertain nonlinear corrupt systems with no precise model under noisy condition. In recent years, the applications of adaptive intelligent robust FSMC have become very common. It is commonly known that the FLS design and stability analysis are not a simple task. The SMC can be used to simplify the design and stability analysis for FLS. On the other hand, the SMC has the inherent chattering phenomena, which can easily be solved by FLS.

### 4.2 Limitations

Most of the hybridisation involving FLC, SMC, FNN, PID, PD and PI to produce a robust, intelligent controller used FLS and SMC employed FLS to approximate the control input $u$ as a nonlinear function of $s$ within a boundary layer. The major disadvantage of this kind of combination is that the value of $\lambda$ has to be predefined by an expert. In addition, this type of combination is intended to approximate only a nonlinear sliding surface excluding nonlinear switching function. The switching function is still linear. This does not have significant contribution to the control performances. This is because the form of the boundary layer does not have much influence on the control performance rather the width of the boundary layer [1]. On the other hand, only few papers, use the information from SMC to design a FLC. Despite the fact that this type of combination cans easily approximates a nonlinear switching function with less rules. However, complicated switching function with less rules can be attained with the resulting FSMC better than classical FLC.

Most of the researchers applied their proposed controller only on simulations. Very few researchers used their proposed controller in real-world environment. It is expected that the simulation result should be validated with experiment in the real-life settings. Another major limitation of FSMC is in choosing the suitable sliding surface. Success was reported for using the PID, PD and PI like sliding surfaces in literatures [28-36]. However, it is difficult to declare one of these sliding surfaces as the best. This is because success was reported for all of them and based on this review there is no study that compares their performances.

Many studies are not after the performance comparison of their proposed controller with the state-of-the-art controllers. This is difficult or even impossible to measure the efficiency and robustness of the proposed controller over the state-of-the-art controllers [75-78].

### 4.3 Future Research

Research effort should be spent on balancing the combination that can positively influence the nonlinear switching function as well as nonlinear switching surface with few rules. Also, the resulting controller should not be complicated. In the future, researchers are recommended to report the performance of different types of sliding surface before choosing the most suitable for use in designing their controllers. We encourage researchers to propose new sliding surface with less complication and better performance than the existing once.

### 5. CONCLUSIONS

The aim of this paper is to review the recent progress on the design of intelligent controller based on sliding mode and fuzzy logic systems. In this paper, we present a comprehensive review of the state of the art studies conducted by researchers designing controllers. The review has shown that researchers heavily depend on the fuzzy sliding mode controller.